\documentclass[letterpaper, 10pt, conference]{ieeeconf}

\IEEEoverridecommandlockouts 

\usepackage{graphicx}
\usepackage{fancyhdr}
\usepackage{amsmath}
\usepackage{amssymb}
\usepackage{booktabs}
\usepackage{array}
\usepackage{collcell}
\usepackage{multirow}
\usepackage{subcaption}
\usepackage{hyperref}
\hypersetup{hidelinks}
\usepackage{xcolor}
\usepackage{url}
\usepackage{listings}
\usepackage{float}
\usepackage{placeins}
\usepackage{pifont}
\usepackage[capitalise]{cleveref}
\usepackage{comment}

\lstdefinestyle{yaml}{ basicstyle=\ttfamily\fontsize{6.2}{7.2}\selectfont, columns=fixed,
 backgroundcolor=\color{gray!10}, frame=none, breaklines=true,
 showstringspaces=false, xleftmargin=1mm, xrightmargin=1mm,
 keepspaces=true, captionpos=b }

\lstdefinestyle{pseudo}{ basicstyle=\ttfamily\fontsize{7.2}{8.2}\selectfont, columns=fullflexible,
 backgroundcolor=\color{gray!8}, frame=none, breaklines=true,
 showstringspaces=false, xleftmargin=1mm, xrightmargin=1mm,
 keepspaces=true, captionpos=b }

\bstctlcite{BSTcontrol}

\newcommand{\sk}[1]{\textcolor{red}{[SK: #1]}}
\newcommand{\fail}{\textcolor{red}{$\times$}}  

\newcommand{\code}[1]{{\setlength{\fboxsep}{1.5pt}\colorbox{gray!10}{\texttt{#1}}}}
\ExplSyntaxOn
\NewDocumentCommand{\slamname}{m}{
  \str_case:nnF {#1} {
    {orbslam3i}{\shortstack[l]{ORB-SLAM3-I}}
    {orbslam3}{\shortstack[l]{ORB-SLAM3}}
    {okvis2x}{OKVIS2-X}
    {cuvslam}{cuVSLAM}
    {droidslam}{DROID-SLAM}
    {dpvslam}{DPV-SLAM}
    {mast3rslam}{MASt3R-SLAM}
    {vggtslam}{VGGT-SLAM}
    {gigaslam}{GigaSLAM}
    {s3pogs}{S3PO-GS}
  } {#1}
}
\ExplSyntaxOff
\newcolumntype{L}{>{\collectcell\slamname}l<{\endcollectcell}}
\definecolor{lossC}{HTML}{B35900}
\definecolor{ateC}{HTML}{1F4E9E}
\definecolor{speC}{HTML}{1B7340}
\definecolor{poseC}{HTML}{666666}
\definecolor{workC}{HTML}{6A3D9A}
\makeatletter
\newcommand{\rescell}[1]{%
  \in@{[}{#1}%
  \ifin@\expandafter\@firstoftwo\else\expandafter\@secondoftwo\fi
  {\res@data#1\res@stop}%
  {\in@{\fail}{#1}\ifin@\in@{/}{#1}\fi
   \ifin@\expandafter\@firstoftwo\else\expandafter\@secondoftwo\fi
   {\res@failed#1\res@stop}{#1}}}
\def\res@data#1[#2]#3\res@stop{\res@split#1\res@mid{#2}{#3}}
\def\res@split#1/#2\res@mid#3#4{%
  \in@{/}{#2}%
  \ifin@\expandafter\@firstoftwo\else\expandafter\@secondoftwo\fi
  {\res@three{#1}#2\res@mid{#3}{#4}}%
  {{\color{ateC}#1}/{\color{speC}#2}\res@tail{#3}{#4}}}
\def\res@three#1#2/#3\res@mid#4#5{%
  {\color{lossC}#1}/{\color{ateC}#2}/{\color{speC}#3}\res@tail{#4}{#5}}
\def\res@tail#1#2{{\color{poseC}[#1]}{\color{workC}\relax#2}}
\def\res@failed#1/#2\res@stop{{\color{lossC}#1}/#2}
\makeatother
\newcolumntype{E}{>{\collectcell\rescell}l<{\endcollectcell}}

\title{\LARGE \bf SLAMSqueezeBench: Comparing SLAM Systems\\
under Resource Constraints}

\author{Mohamed Hefny$^{1}$, Karthik Dantu$^{2}$, Steven Y. Ko$^{1}$%
\thanks{$^{1}$Simon Fraser University. \{mohamed\_hefny, steveyko\}@sfu.ca}%
\thanks{$^{2}$University at Buffalo. kdantu@buffalo.edu}%
}

\begin{document}

\maketitle
\pagestyle{empty}
\fancyhf{}
\renewcommand{\headrulewidth}{0pt}
\renewcommand{\footrulewidth}{0pt}
\fancyfoot[C]{\footnotesize This work has been submitted to the IEEE for possible publication. Copyright may be transferred without notice, after which this version may no longer be accessible.}
\thispagestyle{fancy}
\bstctlcite{BSTcontrol}

\begin{abstract}
Simultaneous localization and mapping (SLAM) is one of the services running on an autonomous robot. 
It is typically run to assist other tasks such as planning, manipulation, etc. All these tasks are
run on edge hardware and are subject to severe resource constraints. However, most SLAM systems
are built and tested in isolation, and their performance is reported as if they are the only task running
on a system. We observe that existing benchmarks lack a common mechanism for comparing SLAM
systems under realistic resource constraints. To address this limitation,
we have developed SLAMSqueezeBench, a framework that allows testing of SLAM systems
under realistic workloads on edge hardware. It does so by imposing constraints on compute and memory resources available for the SLAM system during execution. 
It also simulates realistic camera frame acquisition with frame drops when a finite buffer is full. Using
SLAMSqueezeBench, we compare nine SLAM systems spanning classical systems, learning-based systems, and
approaches for Gaussian splatting. Our testing framework will be available for use by the community upon publication.

\end{abstract}

\section{Introduction}
\label{sec:intro}



Autonomous robots rely on several components running together, including perception, localization, mapping, planning, and control. Simultaneous localization and mapping (SLAM) is one of these components: it estimates the robot's pose while building a map of its surroundings~\cite{cadena2016slam}. 

To support navigation, SLAM must provide pose estimates continuously and keep pace with incoming sensor measurements. For camera input at 25--30~Hz, a new frame arrives roughly every 33--40~ms. The camera continues producing frames regardless of whether SLAM has finished processing earlier ones. Meanwhile, the autonomy software stack runs on embedded hardware, where SLAM shares limited compute and memory with other components. These resource demands can make continuous operation difficult and leave less capacity for applications that depend on SLAM~\cite{semenova2024systems}. Practical SLAM performance therefore depends on whether the system can provide timely, accurate estimates within a shared resource budget.

SLAM systems use different approaches to estimate motion and represent the environment, resulting in different computational demands. Classical geometric systems, such as ORB-SLAM2~\cite{murartal2017orbslam2} and ORB-SLAM3~\cite{campos2021orbslam3}, estimate motion and scene structure from geometric relationships between sensor observations. Learning-based systems incorporate neural networks into tracking and mapping: NeRF-SLAM uses neural radiance fields for dense reconstruction~\cite{rosinol2023nerf}, while UncLe-SLAM learns sensor uncertainty to guide tracking and mapping~\cite{sandstrom2023uncle}. Gaussian-splatting systems, such as SplaTAM~\cite{keetha2024splatam}, represent the scene with three-dimensional Gaussians and optimize camera poses and map parameters through differentiable rendering. These approaches differ in their use of CPU, GPU, and memory. Choosing a system for an autonomous robot therefore requires understanding its performance under different levels of resource availability.

Existing benchmarks establish common evaluation procedures, but do not provide a holistic analysis of SLAM performance under the operating conditions of an autonomy stack. The TUM RGB-D benchmark provides recorded sequences, ground-truth trajectories, and tools to measure trajectory error~\cite{sturm2012evaluating}. Robustness studies extend the evaluation to challenging inputs and  perturbations~\cite{bujanca2021robust,hefny2026sal}. These evaluations help characterize accuracy and robustness, but do not establish whether a system can sustain pose estimation while sharing limited computing resources with other programs. In addition, replaying a sequence by waiting for each frame to finish before supplying the next allows a slow system to still process every frame. Such an evaluation does not expose potential tracking interruptions that can occur when the camera delivers frames continuously regardless of the processing speed, resulting in frame drops.

A holistic performance analysis and evaluation must connect resource availability to the timeliness, continuity, and accuracy of pose estimation. It should be able to constrain CPU, memory, and GPU resources to examine the budgets available on embedded hardware, and introduce competing workloads to capture resource sharing within the autonomy stack. Sensor measurements must continue arriving at their configured rate, with finite buffers that discard frames when processing falls behind. Evaluation can then reveal how resource constraints affect frame loss, whether the system continues producing poses, and whether it recovers after a temporary constraint is removed. Accuracy must also be measured over the full
sequence: a system that reports a few accurate poses and then stops should not appear equivalent to one that provides accurate estimates throughout the run.

We address these requirements through two contributions.
%
First, we design \textbf{\emph{SLAMSqueezeBench}}, a framework for evaluating SLAM
systems under resource constraints. It applies resource caps and competing
workloads through common controls, and delivers frames at a configured rate with
a finite buffer and explicit frame-dropping policies. Constraints can remain
fixed or change during a run, enabling evaluation of both sustained resource
scarcity and temporary contention.
Second, we present \textbf{\emph{a comparative analysis of nine SLAM systems}} spanning
three classical geometric, four learning-based, and two Gaussian-splatting
approaches. We evaluate frame loss, continued pose production, and trajectory
accuracy under resource caps and competing workloads. To complement Absolute
Trajectory Error (ATE), which evaluates only matched pose estimates, we
introduce Stale Pose Error (SPE), which evaluates position error across the full
sequence by holding the latest estimated position between pose timestamps.

We report fourteen findings in \Cref{sec:eval} and highlight three of them
here. First, a processor cap and a competing workload that leave a system the
same average processor use can still produce different outcomes. Second, 
with resource constraints similar to an embedded board, six processor cores with 4\,GiB of memory
and 4\,GiB of GPU memory, only four of the nine systems keep producing pose
estimates: the three classical systems and DPV-SLAM~\cite{lipson2024dpvslam}.
Third, a temporary processor cap can erase what a system estimated before it,
so its trajectory covers only the frames after the cap.

\section{Related Work}
\label{sec:related}

\noindent SLAM benchmarking frameworks provide common procedures for running
SLAM systems and evaluating the results. SLAMBench~\cite{nardi2015slambench}
measures pose accuracy, runtime, and energy use, while
SLAMBench2~\cite{bodin2018slambench2} extends this evaluation to multiple SLAM
systems and adds memory-use measurements. SLAMBench 3.0~\cite{8794369} adds
support for evaluating scene understanding and reconstruction in non-rigid
environments. VSLAM-LAB~\cite{fontan2025vslamlab} unifies building, configuring,
running, and evaluating visual SLAM systems. These frameworks standardize
execution and evaluation, but they do not provide a common mechanism for
applying both resource caps and competing workloads across SLAM systems. Our
study uses SLAMSqueezeBench to apply both kinds of resource constraint and
compare their effects on frame loss, the continued production of pose
estimates, and pose accuracy. To account for frame loss during these
experiments, the framework simulates camera frames arriving at a user-selected
rate and discards frames when a finite buffer fills.

Other work examines the robustness of SLAM systems to challenging sensor
inputs. Datasets such as Oxford RobotCar~\cite{maddern2017robotcar},
4Seasons~\cite{wenzel2020fourseasons}, TartanAir~\cite{wang2020tartanair}, and
SubT-MRS~\cite{zhao2024subtmrs} cover variations in weather, illumination, and
viewpoint. Perturbation frameworks modify recorded sensor data in controlled
ways to evaluate how specific changes in the input affect SLAM
systems~\cite{bujanca2021robust,hefny2026sal,radulov2024slamfuse,xu2025robustego3d}.
Our study instead examines how SLAM systems respond to limits on the processing
capacity and memory available to process camera frames. We impose these
resource constraints through resource caps and competing workloads.

Another line of work adapts SLAM systems to operate with limited processing
capacity and memory. RTAB-Map limits the map data kept in working memory to
reduce the time needed for map updates~\cite{labbe2024rtabmap}.
AdaptSLAM~\cite{chen2023adaptslam} selects keyframes for processing on a mobile
device and an edge server according to the available processing capacity and
communication bandwidth. SPAQ-DL-SLAM~\cite{pudasaini2024spaq} prunes and
quantizes the neural networks in DROID-SLAM~\cite{teed2021droidslam} to reduce
computation and model size. Our study complements system-specific adaptation
by comparing how different SLAM systems respond to resource caps and competing
workloads applied through a common evaluation framework.

A closely related study examines how hardware resources affect frame
processing and pose accuracy by evaluating three classical SLAM systems on a
laptop and an NVIDIA Jetson TX2~\cite{semenova2024systems}. The study shows how processing delays and
synchronization between SLAM modules can lead to frame loss and affect pose
accuracy. However, the two devices differ in processing capacity, memory, and
processor architecture, so comparing the devices does not isolate the effect
of an individual resource. Our study keeps the hardware fixed and varies
resource caps individually and in combination. We also introduce competing
workloads to examine how other programs sharing the hardware affect SLAM
systems. We also change constraints during a run to examine recovery after
resources become available again. Our comparison covers nine SLAM systems
spanning classical geometric, learning-based, and Gaussian-splatting methods.

\section{SLAMSqueezeBench}
\label{sec:framework}

\begin{figure}[t]
\centering
\includegraphics[width=\columnwidth]{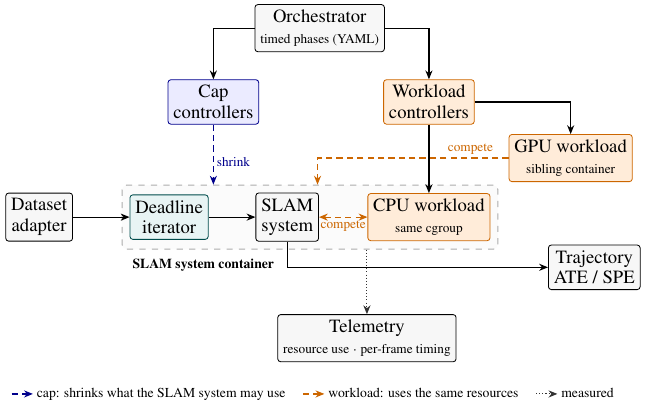}
\caption{SLAMSqueezeBench data flow.}
\label{fig:overview}
\vspace{-2em}
\end{figure}

\noindent SLAMSqueezeBench runs a SLAM system under resource constraints, delivers the
dataset's frames to it at a configured rate, records the resources it uses, and
evaluates the trajectory it produces. It has six main components, shown in \Cref{fig:overview}.
Cap controllers limit the resources available to the SLAM system, and workload
controllers start and stop programs that use the same resources as the SLAM system. The
orchestrator instructs both kinds of controller when to apply, change, or remove
a constraint, which can therefore stay fixed for the whole run or follow a
predefined schedule. The dataset adapter supplies the recorded camera frames in
time order. The deadline iterator delivers them to the SLAM system at the
configured rate and discards frames when a finite buffer fills. Telemetry
records resource use throughout the run. Each run produces a trajectory, from which we compute trajectory errors.
The rest of the section further describes the framework.

\subsection{Resource Caps}
\label{sec:axes}



\noindent 
SLAMSqueezeBench uses resource caps to examine the effects of limited resources
without changing the hardware. The caps limit the resources available to the
SLAM system, approximating some of the restrictions of a smaller machine. A
capped run therefore differs from an uncapped one only in the amount of the
limited resource that the system can use.

SLAMSqueezeBench supports five types of resource cap: CPU,
system memory, disk I/O throughput, GPU memory, and GPU compute.
\Cref{tab:axes} lists each cap, the mechanism that enforces it, and the unit in
which it is set.

We implement the caps with two mechanisms: Linux control groups
(cgroup~v2)~\cite{cgroupv2} for CPU, memory, and disk, and an intercepting
library for GPU. Each SLAM system runs in a container managed by
Podman~\cite{podman}, which allows the framework to configure the CPU, system
memory, and disk I/O caps. Control groups cannot limit GPU memory or compute.
Thus, the framework instead uses HAMi-core~\cite{hamicore}, a library that
intercepts CUDA memory-allocation and kernel-launch calls. Both mechanisms apply
in the same way to every SLAM system.


\begin{table}[t]
\vspace{1em}
\caption{Resource caps supported by SLAMSqueezeBench.}
\label{tab:axes}
\centering
\scriptsize
\setlength{\tabcolsep}{4pt}
\begin{tabular}{@{}llll@{}}
\toprule
Resource & Enforced by & Unit & Models \\
\midrule
CPU & cgroup \code{cpu.max} & cores & shared cores \\
Memory & cgroup \code{memory.max} & MB & fixed memory budget \\
Disk & cgroup \code{io.max} & MB/s & slow or shared disk \\
GPU memory & HAMi-core allocation & MB & smaller or shared GPU \\
GPU compute & HAMi-core launch & \% & shared GPU time \\
\bottomrule
\end{tabular}
\vspace{-3em}
\end{table}

When the SLAM system's use of a resource reaches its cap, one of two things
happens: the caps on CPU time, disk throughput, and GPU compute delay the system,
and the two on memory refuse further allocations. The CPU cap is the control
group's limit on processor time per window~\cite{cgroupv2}: once the SLAM
system's processes exhaust the allowance, the kernel pauses them until the
next one.
The disk I/O cap delays reads and writes that would exceed the configured
throughput. The GPU compute cap delays CUDA kernel launches. Memory caps instead
restrict the amount available to the SLAM system. For system memory, the kernel
can terminate a SLAM system process if the limit is reached and memory cannot
be reclaimed. For GPU memory, HAMi-core rejects allocations that would exceed
the limit. A cap of the first kind therefore slows the SLAM system, while one
of the second kind can stop it.

Resource caps give the SLAM system the amount of processor time, memory, and GPU
compute it would have on a smaller machine, but not the way that machine would
deliver them. For example, a CPU cap sets how much processor time the SLAM
system's processes may use in each 100 ms window. 
The allowance is 100 ms per core, so a one-core cap allows 100 ms and
a two-core cap 200 ms. Under a one-core cap the system can spend that 100 ms in
any combination of cores and time that adds up to it, such as one thread for the
whole window, five threads for 20 ms each, or ten for 10 ms each. The kernel
then pauses its processes until the next window. However, a one-core physical
machine can run only one thread at a time, so the same 100 ms arrives spread
across the whole window rather than in a burst. A memory cap sets how much memory the SLAM system may
allocate, not how quickly it can read and write that memory, which still
happens at the host's speed. A GPU compute cap delays kernel launches, so the SLAM
system gets less time on the GPU rather than a slower one. The framework is extensible, and
controls beyond the ones we implement would bring a capped run closer to a
smaller machine.




\subsection{Competing Workloads}
\label{sec:load}


SLAMSqueezeBench supports two kinds of
competing workloads. Synthetic workloads are programs written specifically to
consume resources through repeated arithmetic or memory operations. Real workloads are applications
that perform tasks a robot might need, such as image segmentation.
SLAMSqueezeBench integrates the SAM 3 segmentation model~\cite{carion2025sam3} as
a real workload.


For synthetic CPU and system-memory workloads, SLAMSqueezeBench uses
\code{stress-ng}~\cite{king2025stressng}, a tool that provides a range of
synthetic workloads. The framework uses three of the available workloads. The
CPU workload uses a configurable number of parallel processes, called workers,
to repeatedly perform double-precision arithmetic. The memory-bandwidth
workload uses a configurable number of workers to repeatedly read and write
large arrays, generating data transfers to and from memory. For the
memory-allocation workload, both the number of workers and the amount of
allocated memory are configurable. The memory-allocation workers repeatedly
allocate and write to memory, reducing the memory available to the SLAM system.
For synthetic GPU workloads, we develop our own GPU workload generator,
SqueezeGPU, in PyTorch. It repeatedly multiplies two
matrices on the GPU, while controlling the GPU utilization of the computation.
Existing generators such as gpu-burn~\cite{gpuburn} and
gpu-fryer~\cite{gpufryer} do not provide this control as they run matrix
multiplications continuously. 

SqueezeGPU provides three configurable settings. The matrix size,
\code{matmul\_n}, determines the amount of computation in each multiplication.
The duty cycle, \code{duty\_cycle}, sets a time budget for repeated matrix
multiplications within each 100\,ms period. For example, a duty cycle of 0.50
targets 50\,ms of repeated multiplications, after which SqueezeGPU sleeps for
the remainder of the period. The GPU-memory setting, \code{vram\_mb}, controls
an additional allocation beyond the memory required for matrix multiplication.
SqueezeGPU runs in a separate container so SLAMSqueezeBench can cap SqueezeGPU's
CPU time and system-memory use without applying those caps to the SLAM system.



\subsection{Real-Time Deadlines}
\label{sec:deadline}

\noindent On a robot, if a SLAM system processes frames slower than the camera
captures them, it could lose frames. Existing benchmarking tools do not
reproduce this loss. The loss occurs between the camera and the SLAM system,
where a finite buffer holds unprocessed frames. 

To reproduce this frame loss behavior, 
SLAM\-Squeeze\-Bench simulates camera frames arriving at a configured
rate with a finite buffer. We implement this frame-delivery
simulation in a component called the deadline iterator. When a new frame
arrives and the simulated buffer is full, the iterator discards a frame
according to the configured drop policy. When the SLAM system requests the
next frame, the iterator supplies the oldest frame from the buffer.
If the buffer is empty, the iterator waits until the next frame arrives.

SLAMSqueezeBench integrates the deadline iterator into the code that supplies
frames to each SLAM system. We provide Python and C++ implementations so the
supported SLAM systems can use the same frame-delivery rules. SLAMSqueezeBench therefore controls frame delivery without a separate interface to robotics middleware such as ROS.

The deadline iterator provides three settings: the buffer capacity, the drop
policy, and the warmup count. The capacity is the number of unprocessed
frames the simulated buffer can hold. When a new frame arrives at a full
buffer, the drop-oldest policy replaces the oldest waiting frame with the new
frame. The drop-newest policy instead discards the new frame and retains the
buffered frames. The warmup count is the number of initial frames the iterator
delivers immediately on request and never drops, so that one-time
initialization delays cause no frame loss. The iterator starts
simulating frame arrivals at the configured rate when the SLAM system requests
the first frame after warmup. 

\subsection{Error Under Dropped Frames}
\label{sec:error}
Frame loss could leave gaps in SLAM pose estimates. Absolute Trajectory Error (ATE)~\cite{sturm2012benchmark}
computes error only over the poses matched with ground truth. For example, a SLAM
system could produce accurate poses for the first few frames and no poses for
the rest of the sequence. ATE would still be low because only the initial
poses are evaluated.

To include the periods without pose estimates, we introduce Stale Pose Error
(SPE). SPE evaluates the position error at every ground-truth timestamp, using
the latest pose estimate available at that time. After each pose estimate, SPE
therefore keeps using it until the next one arrives, or until the end of the
selected sequence if the SLAM system reports no further poses. Because the
robot keeps moving while that estimate stays fixed and goes stale, the error
grows with the distance travelled in that time.

Before the first pose estimate, however, SPE has no previous estimated position
to use. We therefore compare every ground-truth position in this initial
period with the first ground-truth position of the selected sequence. This
convention assumes a known starting position for evaluation, not a pose
reported by the SLAM system. The position error is initially zero and then
reflects the distance from the starting position until the first pose
timestamp.


We compute SPE as the root mean square of the position errors over the selected
sequence. The calculation includes every ground-truth timestamp between the
first and last camera frames selected for the experiment. 
Before calculating
the position errors, we align the estimated trajectory with ground truth using
the same alignment as for ATE. Let $e_i$ denote the position error at the $i$th
ground-truth timestamp and $N$ the number of ground-truth samples in the
evaluation interval. SPE is then $\sqrt{(1/N)\sum_{i=1}^{N} e_{i}^{2}}$.


SPE complements ATE rather than replacing it. It evaluates position errors
during periods without new pose estimates, which ATE ignores. However, some
SLAM systems normally save poses only for a small number of frames, even without
resource constraints. The gaps between saved poses can then lead to high SPE
even when the saved poses are accurate. A high SPE alone therefore does not
establish that the reported poses are inaccurate. We report ATE to assess the
accuracy of the reported poses and SPE to assess position error across the
full selected sequence.

\subsection{Scenarios and Orchestration}
\label{sec:orchestrator}

\noindent SLAMSqueezeBench can apply, change, and remove resource constraints at set
points in a run. On a robot, for example, other programs start and stop, and
the resources left to the SLAM system vary with them. We define these points in
a scenario, a YAML configuration that divides the run into phases, each with
its own constraints. Each scenario uses either phase durations or frame indices in the
input sequence selected for the SLAM system. At each phase transition, the
orchestrator instructs the cap and workload controllers to apply the new
phase's settings. Caps omitted from the next phase are removed, while
workloads omitted from the next phase are stopped or paused.




Throughout the run, the orchestrator records CPU use, system-memory use, and
GPU-memory use at a configurable interval. These measurements help interpret changes in frame
processing and pose accuracy under the caps and workloads.


\subsection{Extensible Benchmarking Architecture}
\label{sec:overview}

\noindent SLAMSqueezeBench is designed to support comparisons beyond the SLAM
systems, datasets, and resource constraints evaluated in this study. The
framework separates dataset handling, SLAM execution, and resource control
through three common interfaces. Therefore, the framework makes a SLAM system,
dataset, or constraint modular and pluggable.

\section{Comparative Analysis}
\label{sec:eval}

\noindent We use SLAMSqueezeBench to compare how resource constraints affect nine SLAM
systems: three classical geometric, four learning-based, and two
Gaussian-splatting. Our evaluation consists of six experiments, E1 to E6. We
measure frame loss and the number of reported pose estimates, and report ATE
and SPE.
E1 progressively tightens a CPU cap. E2 compares CPU caps with competing CPU
workloads that leave a system the same average CPU use. E3 runs synthetic GPU
workloads alongside each system, and E4 replaces them with a real segmentation
model. E5 caps the processor, memory, and GPU memory together, and E6 applies a
temporary CPU cap to ORB-SLAM3 and examines whether it recovers once the cap is
removed.

\subsection{Setup}
\label{sec:setup}

\begin{figure}[t]
\centering
\includegraphics[width=\columnwidth]{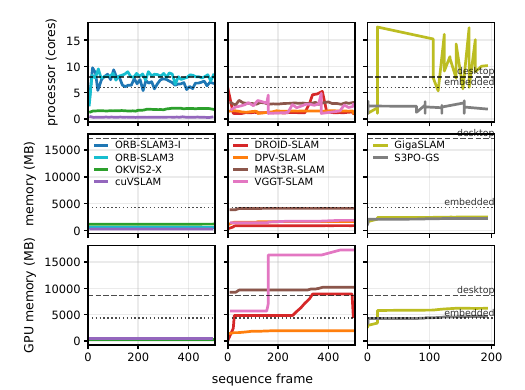}
\caption{Resource use in E1's uncapped runs with the frame-delivery deadline.}
\label{fig:consumption}
\vspace{-2em}
\end{figure}

\noindent We run all experiments on one desktop with an Intel i7-12700K, 12
cores and 20 logical processors, 62\,GiB of memory, and an NVIDIA RTX 3090 with
24\,GiB of GPU memory. Each family uses a dataset that all of its systems
support, EuRoC~\cite{burri2016euroc} for the classical family, TUM
RGB-D~\cite{sturm2012benchmark} for the learning-based one, and
KITTI~\cite{Geiger2012CVPR} for the Gaussian-splatting one. \Cref{tab:setup}
lists each family's sequence, frame count, and delivery rate, which is the
dataset's documented nominal recording rate and stays fixed across
resource-constraint settings. We run ORB-SLAM3 in stereo with and without IMU
input, listed in the result tables as ORB-SLAM3-I and ORB-SLAM3. The deadline
iterator uses a two-frame buffer, the drop-oldest policy, and a warmup count of
zero.
We repeat each system--setting combination three times and identify
unsuccessful runs using execution errors and system-specific tracking-failure
indicators, among other checks. We report the mean percentage of frames lost
over all runs and the mean ATE, SPE, and pose count over successful runs. A
cross in the tables marks a failed setting. \Cref{fig:consumption} shows CPU,
memory, and GPU-memory use in the uncapped runs with the deadline enabled.

\begin{table*}[t]
\vspace{1em}
\caption{Evaluation sequences and configured frame-delivery rates for each
family of SLAM systems.}
\label{tab:setup}
\centering
\scriptsize
\setlength{\tabcolsep}{4pt}
\begin{tabular}{@{}lllrr@{}}
\toprule
Family & Systems & Sequence & Frames & Rate (Hz) \\
\midrule
Classical & ORB-SLAM3 (with/without IMU), OKVIS2-X~\cite{boche2025okvis2x}, cuVSLAM~\cite{korovko2025cuvslam} & EuRoC V1\_01\_easy & 500 & 20 \\
Learning & DROID-SLAM, DPV-SLAM, MASt3R-SLAM~\cite{murai2025mast3rslam}, VGGT-SLAM~\cite{maggio2025vggtslam} & TUM freiburg1\_desk & 500 & 30 \\
Gaussian & GigaSLAM~\cite{deng2025gigaslam}, S3PO-GS~\cite{cheng2025s3pogs} & KITTI 07 & 200 & 10 \\
\bottomrule
\end{tabular}
\vspace{-1em}
\end{table*}

\subsection{Compute Scarcity under a Deadline (E1)}
\label{sec:e1}


\noindent E1 examines how CPU resource caps affect frame loss and pose
estimation. We use two uncapped baselines to distinguish the effects of the
deadline from the additional effects of CPU caps. Without the deadline, the
SLAM system can process every selected frame. With the deadline enabled, the
framework simulates frame arrivals at the configured rate and discards frames
when the buffer fills. We then keep the deadline enabled and apply CPU caps
of four, two, one, and half a core in separate runs. \Cref{tab:e1} presents the
two uncapped baselines followed by the four CPU-cap settings.

\noindent\textbf{Finding 1: Several SLAM systems lose frames under the deadline
even without resource caps.}
The ``no deadline'' and ``deadline'' columns of \Cref{tab:e1} show the
effect of enabling the deadline: frame loss is zero for every system without
the deadline, but increases for several systems when the deadline is enabled.
On EuRoC at 20\,Hz, ORB-SLAM3 without IMU loses 15\% of frames, while the other
classical configurations have no frame loss. The learning-based systems lose
16--75\% on TUM RGB-D at 30\,Hz. On KITTI at 10\,Hz, GigaSLAM and S3PO-GS lose
94\% and 97\%, respectively.

\noindent\textbf{Finding 2: CPU caps above average CPU use can still increase
frame loss.}
With the deadline enabled, DROID-SLAM's uncapped CPU use averages less than
the equivalent of two fully used cores. Nevertheless, a four-core cap
increases frame loss from 34\% to 50\% (\Cref{tab:e1}). \Cref{fig:consumption}
shows that DROID-SLAM's CPU use varies during the run, with peaks above five
cores. The CPU cap limits processor time within each 100\,ms window, rather
than only limiting average use across the entire run. Short periods of high
CPU use can therefore exhaust the processor-time allowance and pause the SLAM
system until the next window, even when average CPU use is below the cap.

\noindent\textbf{Finding 3: Tight CPU caps can cause tracking failure, not just
reduce pose output.}
At the half-core cap, OKVIS2-X, cuVSLAM, DROID-SLAM, DPV-SLAM, and VGGT-SLAM
still report pose estimates (\Cref{tab:e1}). For example, OKVIS2-X reports
95 poses compared with 498 in the uncapped deadline baseline, as frame loss
increases from 0\% to 81\%. CPU caps can also disrupt tracking: ORB-SLAM3's
settings at two cores failed, both with and
without IMU. The failed ORB-SLAM3 runs show tracking loss without successful
relocalisation or map resets that discard the current map.

\noindent\textbf{Finding 4: SPE reveals increased position error under CPU caps
that ATE alone misses.}
Across the tested CPU caps, several SLAM systems report fewer poses and higher
SPE despite unchanged or lower ATE (\Cref{tab:e1}). For example, the half-core
cap reduces VGGT-SLAM's reported pose count from 41 to 17 relative to the
uncapped deadline baseline. ATE remains 0.021\,m at 0.5 cores, but SPE increases from 0.313
to 0.858\,m. The increased SPE reveals greater position error across the full
selected sequence, including periods without new pose estimates. ATE does not
reflect this increase because ATE evaluates only the reported poses.

\begin{table*}[t]
\caption{E1: CPU-cap results \& uncapped baseline. An entry shows \textcolor{lossC}{frame loss
(\%)}\,/\,\textcolor{ateC}{ATE (m)}\,/\,\textcolor{speC}{SPE (m)}
\textcolor{poseC}{[$K$]}, where $K$ is the number of reported poses.}
\label{tab:e1}
\centering
\scriptsize
\setlength{\tabcolsep}{3.3pt}
\begin{tabular}{@{}LEEEEEE@{}}
\toprule
& \multicolumn{2}{c}{Baseline} & \multicolumn{4}{c}{CPU cap} \\
\cmidrule(lr){2-3} \cmidrule(lr){4-7}
System & no deadline & deadline & 4 cores & 2 cores & 1 core & 0.5 cores \\
\midrule
orbslam3i & 0/0.028/0.026 [397] & 0/0.027/0.026 [396] & 44/0.026/0.031 [230] & 74 / \fail & 88 / \fail & 94 / \fail \\
orbslam3 & 0/0.072/0.072 [500] & 15/0.073/0.073 [424] & 62/0.192/0.248 [200] & 76 / \fail & 91 / \fail & 94 / \fail \\
okvis2x & 0/0.026/0.027 [499] & 0/0.027/0.028 [498] & 0/0.026/0.027 [497] & 1/0.027/0.028 [494] & 58/0.025/0.038 [211] & 81/0.024/0.088 [95] \\
cuvslam & 0/0.070/0.071 [500] & 0/0.070/0.071 [500] & 0/0.071/0.071 [499] & 0/0.070/0.071 [500] & 0/0.070/0.071 [499] & 0/0.071/0.071 [499] \\
droidslam & 0/0.428/0.480 [57] & 34/0.436/0.484 [57] & 50/0.421/0.486 [54] & 58/0.399/0.496 [47] & 66/0.450/0.773 [40] & 88/0.420/0.747 [36] \\
dpvslam & 0/0.017/0.021 [500] & 16/0.409/0.427 [420] & 16/0.380/0.399 [419] & 16/0.456/0.469 [419] & 34/0.497/0.509 [327] & 79/0.301/0.469 [105] \\
mast3rslam & 0/0.325/0.341 [13] & 75/0.333/0.321 [11] & 75/0.327/0.341 [12] & 90/0.325/0.304 [11] & 96/0.292/0.377 [8] & 99 / \fail \\
vggtslam & 0/0.030/0.181 [50] & 28/0.021/0.313 [41] & 31/0.026/0.302 [34] & 31/0.037/0.314 [34] & 52/0.040/0.663 [25] & 73/0.021/0.858 [17] \\
gigaslam & 0/0.274/0.939 [113] & 94/18.276/20.377 [12] & 98 / \fail & 99 / \fail & 99 / \fail & 99 / \fail \\
s3pogs & 0/1.235/1.419 [63] & 97 / \fail & 97 / \fail & 98 / \fail & 99 / \fail & 99 / \fail \\
\bottomrule
\end{tabular}
\vspace{-1em}
\end{table*}

\subsection{Caps versus Competing Workload (E2)}
\label{sec:e2}

\noindent A CPU cap sets an upper bound on the processor time a SLAM system
can use, whereas a competing CPU workload runs additional processes that use
the same CPU. E2 examines whether CPU caps and competing workloads affect
frame processing and pose estimation differently when the SLAM system's
average CPU use is similar. For each SLAM system, we measure the average CPU
use under each CPU cap in E1. We then run the SLAM system without a CPU cap,
alongside parallel CPU-workload processes called workers. We adjust the
number of workers, aiming to make the SLAM system's average CPU use similar
to the average measured in the capped run.

\noindent\textbf{Finding 1: CPU caps and competing workloads can produce
different tracking outcomes despite similar average CPU use.}
For ORB-SLAM3 with IMU, for example, a one-core cap and a workload of 30 CPU workers both
result in average CPU use of approximately one core. However,
capped runs fail, whereas the workload runs succeed. ORB-SLAM3 without IMU shows a similar
contrast.

\noindent\textbf{Finding 2: Some SLAM systems report fewer poses and higher
errors under competing CPU workloads than under CPU caps.}
For DPV-SLAM, the one-core cap and a workload of 23 CPU workers produce
similar average CPU use. However, DPV-SLAM reports 327 poses under the cap
and 283 under the workload (\Cref{tab:e1,tab:e2}). ATE increases from
0.497\,m under the cap to 0.594\,m under the workload, and SPE increases
from 0.509 to 0.668\,m. Unlike the improved tracking outcome observed for
ORB-SLAM3, DPV-SLAM reports fewer poses and greater position errors under
the competing workload.


\begin{table*}[t]
\vspace{1em}
\caption{E2: Results under competing CPU workloads. Columns identify the
corresponding CPU-cap settings in E1. The \textcolor{workC}{worker count} is in
parentheses.}
\label{tab:e2}
\centering
\scriptsize
\setlength{\tabcolsep}{4pt}
\begin{tabular}{@{}LEEEE@{}}
\toprule
System & 4 cores & 2 cores & 1 core & 0.5 cores \\
\midrule
orbslam3i & 0/0.027/0.026 [396] (0) & 3/0.027/0.026 [382] (18) & 28/0.026/0.030 [271] (30) & 70 / \fail{} (161) \\
orbslam3 & 15/0.073/0.073 [424] (0) & 13/0.072/0.073 [433] (18) & 19/0.106/0.104 [404] (20) & 56/0.163/0.179 [220] (86) \\
okvis2x & 13/0.025/0.029 [436] (10) & 17/0.026/0.030 [415] (16) & 48/0.026/0.036 [260] (19) & 68/0.025/0.059 [159] (63) \\
cuvslam & 0/0.070/0.071 [500] (0) & 0/0.070/0.071 [500] (0) & 0/0.070/0.071 [500] (0) & 0/0.070/0.071 [500] (0) \\
droidslam & 68/0.422/0.481 [54] (22) & 75/0.406/0.480 [51] (32) & 85/0.395/0.499 [44] (64) & 87/0.599/0.739 [32] (162) \\
dpvslam & 33/0.481/0.487 [336] (19) & 27/0.630/0.603 [365] (18) & 43/0.594/0.668 [283] (23) & 88/0.571/0.832 [59] (58) \\
mast3rslam & 90/0.331/0.285 [12] (22) & 93/0.330/0.288 [10] (36) & 95/0.316/0.340 [10] (71) & 97/0.311/0.371 [8] (143) \\
vggtslam & 51/0.032/0.532 [27] (34) & 77/0.049/0.879 [19] (64) & 89/0.020/1.035 [17] (119) & 92/0.019/1.069 [17] (232) \\
gigaslam & 98 / \fail{} (21) & 99 / \fail{} (42) & 99 / \fail{} (75) & 99 / \fail{} (173) \\
s3pogs & 99 / \fail{} (29) & 99 / \fail{} (33) & 99 / \fail{} (53) & 99 / \fail{} (127) \\
\bottomrule
\end{tabular}
\vspace{-1em}
\end{table*}

\subsection{Contention on a Shared GPU (E3)}
\label{sec:e3}

\noindent E3 examines how a competing GPU workload affects frame processing
and pose estimation. We run SqueezeGPU alongside each SLAM system with the
deadline enabled. SqueezeGPU repeatedly multiplies matrices on the GPU, with
a duty cycle that sets the computation time budget within each 100\,ms
period. For example, a duty cycle of 0.50 targets 50\,ms of repeated
multiplications, after which SqueezeGPU sleeps for the remainder of the
period. We test five workload settings. The first four use $4096 \times 4096$
matrices and duty cycles of 0.25, 0.50, 0.75, and 0.95 to examine the effect
of increasing the computation time budget. The fifth uses $8192 \times 8192$
matrices at a duty cycle of 0.95 to examine the effect of larger matrix
multiplications.

\noindent\textbf{Finding 1: cuVSLAM's pose output and accuracy remain nearly
unchanged under competing GPU workloads.}
Unlike ORB-SLAM3 and OKVIS2-X, cuVSLAM performs computation on the GPU.
Nevertheless, cuVSLAM reports 499--500 poses across all five workload
settings, compared with 500 in the uncapped deadline baseline. ATE remains
between 0.070 and 0.071\,m, and SPE remains at 0.071\,m (\Cref{tab:e3}).
This stability does not mean that cuVSLAM provides the most accurate pose
estimates. Under the same workload settings, ORB-SLAM3 with IMU and OKVIS2-X
both achieve lower ATE and SPE, with both metrics between 0.025 and 0.028\,m.

\noindent\textbf{Finding 2: Increasing the GPU workload's duty cycle affects
pose output differently across the learning-based SLAM systems.}
With $4096 \times 4096$ matrices, increasing the duty cycle from 0.25 to
0.95 reduces DROID-SLAM's reported pose count from 61 to 26 and DPV-SLAM's
from 346 to 116 (\Cref{tab:e3}). Frame loss also increases, from 56\% to 93\%
for DROID-SLAM and from 31\% to 77\% for DPV-SLAM. In contrast, VGGT-SLAM
reports 34 poses at all four duty cycles, while MASt3R-SLAM reports 11--12.
The four learning-based systems therefore differ substantially in how much
additional pose output they lose as the competing workload's computation
time budget increases, despite using the same sequence and frame rate.

\noindent\textbf{Finding 3: Larger workload matrices can reduce pose output
even when the duty cycle is unchanged.}
At a duty cycle of 0.95, increasing SqueezeGPU's matrix size from
$4096 \times 4096$ to $8192 \times 8192$ reduces DROID-SLAM's reported pose
count from 26 to 20 and DPV-SLAM's from 116 to 101 (\Cref{tab:e3}).
The configured duty cycle alone therefore does not fully describe how a
competing GPU workload affects a SLAM system.

\begin{table*}[t]
\begin{minipage}[b]{0.745\textwidth}
\caption{E3: Results under synthetic GPU workloads.}
\label{tab:e3}
\centering
\scriptsize
\setlength{\tabcolsep}{2.5pt}
\begin{tabular}{@{}LEEEEE@{}}
\toprule
& \multicolumn{4}{c}{Duty cycle, matrix size 4096} & Matrix size 8192 \\
\cmidrule(lr){2-5} \cmidrule(lr){6-6}
System & 0.25 & 0.50 & 0.75 & 0.95 & duty 0.95 \\
\midrule
orbslam3i & 0/0.026/0.025 [397] & 0/0.027/0.026 [397] & 0/0.027/0.026 [397] & 0/0.027/0.026 [397] & 0/0.027/0.026 [397] \\
orbslam3 & 15/0.073/0.074 [427] & 13/0.072/0.072 [437] & 7/0.073/0.073 [464] & 6/0.070/0.071 [468] & 7/0.119/0.120 [466] \\
okvis2x & 0/0.027/0.028 [497] & 0/0.025/0.026 [497] & 1/0.026/0.028 [494] & 0/0.026/0.027 [499] & 0/0.026/0.027 [498] \\
cuvslam & 0/0.071/0.071 [499] & 0/0.070/0.071 [500] & 0/0.070/0.071 [500] & 0/0.070/0.071 [500] & 0/0.070/0.071 [500] \\
droidslam & 56/0.429/0.487 [61] & 76/0.435/0.467 [53] & 89/0.430/0.488 [44] & 93/0.557/0.573 [26] & 94/0.739/0.754 [20] \\
dpvslam & 31/0.327/0.339 [346] & 48/0.385/0.380 [261] & 67/0.554/0.563 [166] & 77/0.475/0.491 [116] & 80/0.635/0.631 [101] \\
mast3rslam & 80/0.333/0.289 [12] & 84/0.335/0.301 [12] & 87/0.325/0.290 [11] & 90/0.317/0.262 [12] & 90/0.323/0.281 [11] \\
vggtslam & 31/0.021/0.297 [34] & 31/0.032/0.301 [34] & 32/0.030/0.311 [34] & 34/0.024/0.321 [34] & 34/0.024/0.335 [33] \\
gigaslam & 96/22.976/28.413 [8] & 99 / \fail & 99 / \fail & 99 / \fail & 99 / \fail \\
s3pogs & 98 / \fail & 98 / \fail & 99 / \fail & 99 / \fail & 99 / \fail \\
\bottomrule
\end{tabular}
\vspace{-2em}
\end{minipage}%

\hfill
\begin{minipage}[b]{0.235\textwidth}
\hyphenpenalty=10000 
\caption{E4: Results with SAM 3 running alongside SLAM systems.}
\label{tab:e4}
\centering
\scriptsize
\setlength{\tabcolsep}{4pt}
\begin{tabular}{@{}LE@{}}
\toprule
System & Beside the co-tenant \\
\midrule
orbslam3i & 0/0.027/0.026 [396] \\
orbslam3 & 6/0.072/0.072 [469] \\
okvis2x & 0/0.026/0.027 [497] \\
cuvslam & 0/0.070/0.071 [500] \\
droidslam & 92/0.452/0.507 [33] \\
dpvslam & 71/0.569/0.608 [144] \\
mast3rslam & 87/0.329/0.269 [13] \\
vggtslam & 32/0.024/0.313 [34] \\
gigaslam & 99 / \fail \\
s3pogs & 99 / \fail \\
\bottomrule
\end{tabular}
\vspace{-2em}
\end{minipage}%

\end{table*}

\subsection{A Real Co-Tenant (E4)}
\label{sec:e4}

\noindent E4 examines how a real competing workload, rather than a synthetic one,
affects frame processing and pose estimation. We run SAM~3~\cite{carion2025sam3},
a segmentation model that takes a word and finds it in an image, alongside each
SLAM system with the deadline enabled. SAM~3 reads the same camera frames as the
SLAM system on the same GPU, which resembles the arrangement on a robot, where
one camera publishes frames and several programs read them. We give it a single word for all three datasets, and when it reaches the last frame of the
sequence it starts again from the first, so it runs for as long as the SLAM
system does. Unlike the synthetic workload in E3, SAM~3 has no duty cycle
setting: it uses as much of the GPU as it needs.

\noindent\textbf{Finding 1: The co-tenant does not affect the systems that do
little or no computation on the GPU.}
ORB-SLAM3 and OKVIS2-X do no computation on the GPU, and cuVSLAM keeps it busy
less than one per cent of the time on its own. Running beside SAM~3, these
three lose no more frames than they do without it. Their ATE and SPE in the
first four rows of \Cref{tab:e4} match the deadline column of \Cref{tab:e1} to
within 0.001\,m.

\noindent\textbf{Finding 2: For the systems that compute on the GPU, the
co-tenant has the same effect as a synthetic workload with the same share of
the GPU.}
SAM~3 keeps the GPU busy 77\% of the time. With SAM~3 running, each of the six
GPU systems loses about as many frames as it lost in E3 at duty cycles of 0.75
and 0.95, which occupy the GPU 75\% and 95\% of the time. DPV-SLAM, for
example, loses 71\% of its frames beside SAM~3 and 67\% and 77\% at the two
duty cycles, and reports 144 poses (\Cref{tab:e4}) against 166 and 116
(\Cref{tab:e3}). A real co-tenant therefore affects a SLAM system mainly through
the share of the GPU it takes, and a synthetic workload with a similar duty
cycle approximates its effect.

\subsection{Which Machines Can Run Each System (E5)}
\label{sec:e5}

\noindent E5 examines whether each system runs as expected on a smaller machine, where
several resources are short at once. E1 to E4 constrain one resource at a
time. Here we cap the processor, system memory, and GPU memory together, at
two budgets: that of a smaller desktop, eight cores with 16\,GiB of memory and
8\,GiB of GPU memory, and that of an embedded board, six cores with 4\,GiB of
memory and 4\,GiB of GPU memory. We run each system with nothing capped and at
both budgets, with the
deadline enabled. \Cref{fig:consumption} shows what each system uses with
nothing capped, against the two budgets.

\noindent\textbf{Finding 1: A system's memory use predicts whether it runs at a
budget.}
A system runs at a budget when the memory and GPU memory it uses with nothing
capped fit within that budget (\Cref{fig:consumption} and \Cref{tab:e5}).
MASt3R-SLAM and VGGT-SLAM use more GPU memory than the desktop budget allows, so
they fail at both budgets. GigaSLAM exceeds only the embedded budget and fails
only there. The three classical systems and DPV-SLAM fit within both budgets
and run at both. DROID-SLAM is a borderline case: with nothing capped it uses up
to 8.3\,GiB of GPU memory, just over the desktop budget's 8\,GiB, yet it runs at
that budget and fails only at the embedded one. S3PO-GS fails at every budget
for another reason: as E1 shows, it cannot keep up with the camera even with
nothing capped.

\noindent\textbf{Finding 2: Of the classical systems, only ORB-SLAM3 without IMU loses accuracy as the budget shrinks.}
ORB-SLAM3 with IMU, OKVIS2-X, and cuVSLAM report ATE and SPE within 0.002\,m of
their uncapped values at both budgets (rows one, three, and four of
\Cref{tab:e5}). ORB-SLAM3 without IMU keeps running at both budgets, but its
ATE rises from 0.073 to 0.107 and 0.148\,m and its pose count falls from 424 to
333 and 273 across the three columns.


\begin{table}[t]
\caption{E5: Results under combined resource caps. The uncapped column is the
baseline. Column headers show CPU cores\,/\,system memory (GiB)\,/\,GPU memory
(GiB). A cross with no frame loss marks a system that failed before its first
frame.}
\label{tab:e5}
\centering
\scriptsize
\setlength{\tabcolsep}{2pt}
\begin{tabular}{@{}LEEE@{}}
\toprule
& Uncapped & Desktop budget & Embedded budget \\
System & 20 / 62 / 24 & 8 / 16 / 8 & 6 / 4 / 4 \\
\midrule
orbslam3i & 0/0.027/0.026 [396] & 0/0.027/0.025 [397] & 12/0.026/0.026 [373] \\
orbslam3 & 15/0.073/0.073 [424] & 33/0.107/0.111 [333] & 48/0.148/0.164 [273] \\
okvis2x & 0/0.027/0.028 [498] & 1/0.026/0.027 [492] & 1/0.026/0.027 [494] \\
cuvslam & 0/0.070/0.071 [500] & 0/0.070/0.071 [497] & 0/0.071/0.071 [497] \\
droidslam & 34/0.436/0.484 [57] & 36/0.441/0.483 [58] & \fail \\
dpvslam & 16/0.409/0.427 [420] & 16/0.278/0.293 [421] & 16/0.475/0.481 [420] \\
mast3rslam & 75/0.333/0.321 [11] & 71 / \fail & \fail \\
vggtslam & 28/0.021/0.313 [41] & \fail & \fail \\
gigaslam & 94/18.276/20.377 [12] & 97/25.081/32.732 [5] & \fail \\
s3pogs & 97 / \fail & 96 / \fail & \fail \\
\bottomrule
\end{tabular}
\vspace{-3em}
\end{table}

\subsection{Recovery after a Transient Squeeze (E6)}
\label{sec:e6}

\noindent E6 examines whether a SLAM system recovers once a temporary cap is lifted.
The other experiments hold their caps and workloads fixed for the whole run.
Here a run has three phases: a clean phase with nothing applied for the first
30\% of the sequence, a squeeze phase under a half-core cap for the next 20\%,
and a released phase with the cap lifted for the remaining 50\%.
The boundaries are frame indices, so the squeeze covers the same part of the
trajectory in every run. In each phase we measure the
share of frames the system receives and the share of received frames it
produces a pose for. We run this experiment on ORB-SLAM3 only, with and without IMU, and squeeze it to half a core, because \cref{tab:e1} shows it
failing at that cap, so the question is what it does when the cap lifts.


\noindent\textbf{Finding 1: ORB-SLAM3 recovers after the cap lifts, but its saved
trajectory no longer contains the frames from before recovery.}
During the squeeze the system receives only 17 to 26\% of the frames. In five
of the six runs these gaps make it lose tracking and reset its map. The poses
it had estimated in the clean phase belonged to that map and are deleted with
it. After the release it receives 71 to 93\% of the frames again, builds a new
map, and produces a pose for 98 to 100\% of them. Its saved trajectory
therefore starts after the cap is lifted.

\section{Conclusion}
\label{sec:conclusion}

\noindent 
We introduce SLAMSqueezeBench, a framework that imposes resource constraints through caps and competing workloads, delivers frames at a fixed, configured rate, and discards them when a finite buffer fills. Using it to compare nine SLAM systems, we find that tracking and pose
output depend on how and when resource availability is reduced, not only on average resource use, and that resource constraints can affect whether a system continues producing pose estimates and which portions of its saved trajectory remain available. Because ATE evaluates
only reported poses, we introduce Stale Pose Error (SPE) to evaluate position error throughout the selected sequence, including the gaps in pose output. We will release SLAMSqueezeBench at \url{https://github.com/sfu-rsl/SLAMAdversarialLab}.

\bibliographystyle{IEEEtran}
\bibliography{references}

\end{document}